\documentclass[sigconf]{acmart}

\usepackage{multicol}

\AtBeginDocument{ \providecommand\BibTeX{{ \normalfont B\kern-0.5em{\scshape i\kern-0.25em b}\kern-0.8em\TeX}}}

\setcopyright{acmcopyright}
\copyrightyear{2022}
\acmYear{2022}
\setcopyright{rightsretained}
\acmConference[RecSys '22]{Sixteenth ACM Conference on Recommender Systems}{September 18-23, 2022}{Seattle, WA, USA}
\acmBooktitle{Sixteenth ACM Conference on Recommender Systems (RecSys '22), September 18-23, 2022, Seattle, WA,  USA}
\acmDOI{10.1145/3523227.3547383}
\acmISBN{978-1-4503-9278-5/22/09}

\begin{document}

\title{Exploration with Model Uncertainty at Extreme Scale in Real-Time Bidding}

\author{Jan Hartman}
\email{jhartman@outbrain.com}
\affiliation{
  \institution{Zemanta, an Outbrain company}
  \country{Slovenia}
  \city{Ljubljana}
}

\author{Davorin Kopi\v{c}}
\email{dkopic@outbrain.com}
\affiliation{
  \institution{Zemanta, an Outbrain company}
  \country{Slovenia}
  \city{Ljubljana}
}

\begin{abstract}

In this work, we present a scalable and efficient system for exploring the supply landscape in real-time bidding. 
The system directs exploration based on the predictive uncertainty of models used for click-through rate prediction and works in a high-throughput, low-latency environment. 
Through online A/B testing, we demonstrate that exploration with model uncertainty has a positive impact on model performance and business KPIs.
 
\end{abstract}

\ccsdesc[500]{Computing methodologies~Machine learning}
\ccsdesc[500]{Information systems~Computational advertising}

\keywords{machine learning, big data, real-time bidding, exploration, uncertainty}

\maketitle

\section{Introduction}

In this work, we describe a large-scale exploration system that utilizes the predictive uncertainty of machine learning models at Zemanta, an Outbrain company. Zemanta is a demand-side platform (DSP) in the real-time bidding (RTB) ecosystem~\cite{wang2016displayrtb}. In RTB, several DSPs (bidders) compete for advertising space by bidding for it in real-time while a web page is loading. Advertising space is sold on a per-ad impression basis, which enables selling it at market value. Every time a user visits a web page with ad space, DSPs receive a bid request for that ad impression, to which they can respond with an ad and bid price. 
It is important to state that DSPs typically do not receive any information from auctions they did not win (e.g. user feedback or the winning price), which means we are dealing with censored data. 

RTB can be divided into two components: supply (the ad space on publishers' sites), and demand (the ads that can be shown and the budgets associated with them). Exploring both sides efficiently is among the most challenging tasks of recommender systems used in RTB.
One of the more common strategies in RTB is cost-per-click (CPC) bidding, where a bidder wants to buy clicks at a certain cost. This is typically done using machine learning models for click-through rate (CTR) prediction, where the predicted CTR of an ad impression is directly used in determining the amount of money the bidder is willing to pay for it. Having a good click prediction model is thus of significant importance. Many new algorithms and modeling techniques are proposed each year by researchers in academia and industry~\cite{mcmahan2013ad, he2014practical, cheng2016wide, zhou2018deep}. 

The RTB field has a few intrinsic properties: large amounts of data (Zemanta receives millions of bid requests per second) and a low latency requirement (typically around 100 ms). The RTB network connects advertisers to millions of publishers, which makes the market enormous and challenging to tackle in terms of recommendations. The distribution of data also changes rapidly as new publishers, ads, and users enter and exit, so models need to be updated with new data frequently to stay competitive. Zemanta uses deep models from the DeepFM family~\cite{guo2017deepfm} to predict CTR, computing over 600 million predictions per second~\cite{hartman2021scaling}. At this scale, having efficient predictive models and training pipelines is essential, which was an important guideline in designing our exploration system that heavily leverages ML.

\section{Motivation}

Many recommender systems use predictive models trained on past behavior of the system, which means feedback loops are a common pitfall in this area. CTR models used in RTB are no exception here: we use their predictions to set the bid price for ad impressions and then train them on the same impressions that we bought. By doing this, we are essentially creating the models' training data with the models themselves. A model that has not observed certain parts of supply (e.g. publishers) or demand (ads) enough will produce poor predictions on these areas and will therefore buy fewer impressions from those areas, therefore completing the loop. To remediate this, we force the models to explore so they gain more knowledge in areas that they are unfamiliar with. We must also consider that any exploration system we implement must run continuously due to RTB's quickly-changing landscape of publishers, advertisers, and users. 

In this work, we focus on exploring the supply (incoming requests). This task is somewhat different compared to the usual exploration of a recommender system's inventory (ads, videos, products, etc). A DSP can choose not to respond to a bid request and thus disregard a part of supply (e.g. certain publishers) entirely. Conversely, even if the DSP responds, they must still win the auction to get the click feedback data that is used to train CTR models. This situation with censored data makes exploration in this context a demanding task. 

As mentioned above, we want to explore parts of supply where our model is not performing well. To do this, we must define how to actually explore and how to determine the model's performance. The first point is conceptually simple: we explore by increasing our bid price for certain impressions. This means that we are more likely to win and buy these impressions and have the model learn from them. However, this has trade-offs. We are essentially paying extra for additional training data and potentially impacting business KPIs. This means that our exploration has to pay off in the end by improving our model and thus getting better business results. 

There are many options on how to determine the model's performance on certain supply. One simple approach could be counting, e.g. saying that we need $N$ impressions from a certain publisher in order for the model to perform well there~\cite{liu2019bid}. However, counting is problematic as we would need to manually deal with setting thresholds on all possible dimensions of exploration. We wanted an approach that could deal with many dimensions at once and would be tightly correlated to any running CTR prediction model's knowledge. One such direction and a growing area of research is predictive uncertainty. Our idea was to use the uncertainty of a CTR model's predictions to determine where and how much to explore. A higher level of uncertainty should mean the model knows less about certain supply. If our goal is to improve the model, we should focus on buying those uncertain parts of supply.

\section{Implementation}

Our main criteria when surveying methods for predictive uncertainty estimation were their performance and scalability. Due to the massive scale we operate on, any implementations that we want to use must be efficient. 
An overview of such methods can be found in Gawlikowski et al.~\cite{gawlikowski2021survey}, while Ovadia et al.~\cite{ovadia2019can} tested a few more performant methods on dataset shift. 
The method we ultimately chose for our use case was Monte Carlo (MC) dropout~\cite{gal2016dropout}, which is based on the well-known dropout regularization technique for neural networks~\cite{srivastava2014dropout}. However, there exist other well-performing and relatively simple methods~\cite{lakshminarayanan2017simple}. Dropout is based on randomly zeroing out neurons in the network at training time, which prevents overfitting and makes the model more robust. MC dropout works by also applying dropout at prediction time, thus making the predictions slightly randomized. By computing many predictions for a given data instance, we can estimate the predictive distribution. The standard deviation of this sample of predictions represents the uncertainty of the model -- if the predictions vary significantly, it means the model is more uncertain. MC dropout is also easy to integrate as any deep model trained with dropout can use it by merely enabling dropout at prediction time.

Like most recommender system settings where many items are scored and then ranked for each request, we score many ads with our CTR prediction model on each bid request we receive, i.e. we compute one prediction per ad. Most methods for uncertainty estimation require much more computation than making a prediction. For example, MC dropout requires making $N$ predictions, which means we would be computing $\#ads \times N$ instead of $\#ads$ CTR predictions. Because we use deep models, this would mean a substantial increase in compute usage. This was one of the reasons we decided to explore with uncertainty only on the supply side -- instead of making $N$ times the predictions (total  $\#ads \times N$), we can make only $N$ additional predictions per request (total  $\#ads + N$), which is much more scalable. Apart from being more efficient, this approach also enables us to tackle supply and demand exploration separately.
This approach also brings another consideration: if we have a CTR model that uses both ad and request data, how do we make predictions only on the request level? We opted to solve this with a simple masking approach, where all ad features are replaced with dummy constants, leaving only request features to be used in the uncertainty computation.

In the previous section, we mentioned that we want our models to explore the supply where they lack knowledge. This is equivalent to exploring where the estimated predictive uncertainty is high, which in our case means increasing our bid price. Consequently, we have to define a transformation from uncertainty into a bid modifier. As we are directly impacting the business with exploration, this transformation must be robust to prevent spending too much on exploration. Moreover, it must adjust continuously due to new and unpredictable incoming traffic.
For this purpose, we implemented a lightweight controller. The controller is based on streaming (Apache Kafka~\cite{kreps2011kafka}). It continuously receives computed uncertainties and adjusts a handful of statistic estimates used by our bidder when determining how much to explore. These uncertainty statistics are computed across a few dimensions, which helps with more accurate estimates.
We calculate the bid modifier based on the mean uncertainty in the past few minutes: $modifier = unc. / \mu_{unc.}$ while using a few quantile values for filtering requests with either too low or too high uncertainty. This system works in near real-time and automatically adjusts the obtained modifiers to enable continuous exploration without compromising business goals. It has a few parameters that can adjust the level of exploration (explore-exploit ratio): the percentage of requests to explore, the uncertainty thresholds, and the minimum and maximum bid modifier that can be applied.

\section{Evaluation}

Our main way of evaluating our approach was to conduct A/B tests in a live production environment. The test groups were identical except for the CTR models and exploration strategies. Each group had a separate CTR model, all models had identical feature sets, hyperparameters, and starting weights. However, during the test, each of the models only trained on the data it bought. This prevents data leakage, as a model could otherwise exploit another model's explored impressions. In this way, we could judge if the models had improved through exploration. 
We set up three groups: 
\begin{itemize}
    \item \textbf{Control group}: the default, it does not perform any explicit supply exploration.
    \item \textbf{Uncertainty explore group}: explores with the uncertainty mechanism described above.
    \item \textbf{Random explore group}: explores by increasing the bid price with a random modifier. This modifier is sampled from the distribution of the modifier obtained with uncertainty.
\end{itemize}

The goal was to compare our approach with a non-exploratory strategy and one that explores randomly. This helps to confirm the idea of exploring by increasing the bid price, along with the validity of using uncertainty for exploration. We matched the distributions of bid increases for the two explore groups to make sure they both explore equally.

The online part of our tests was conducted on a subset of incoming traffic, where each model received an equal amount of requests, distributed randomly. We ran the test for several days, where each model bought hundreds of millions of impressions. Our KPIs here were revenue and CTR. Note that revenue can be easily increased by just blindly increasing bids and buying more, but this would lead to not satisfying business constraints, i.e. buying at CPC prices higher than defined. We report revenue increases where business constraints were respected.
After we concluded the online test, we used the last state of the online models and tested them offline on data we obtained after this time (the models had not seen this data). Our goal was to measure whether the exploration strategies improved the models, so our metrics of choice were AUC and log loss.

The results are displayed in Table~\ref{tab:results}. 
We can observe that both strategies show improvements over the control group in revenue and AUC, but not CTR and log loss. The uncertainty-based strategy also outperforms the random one in all metrics. This confirms our hypotheses that using the model's uncertainty for exploration is beneficial overall and that it is better than exploring randomly. The random exploration strategy obtained a surprising lift in revenue, but the decrease in CTR along with the worse log loss show that it does not manage to improve the model. The results also show that apart from business KPIs, the uncertainty exploration improves the model as well, as the AUC and log loss are better. In the long run, this should lead to even better results. We also computed model uncertainties on the offline dataset and observed that the uncertainty explore model's uncertainty was on average 14\% lower than the uncertainty of the random explore model, which further indicates that it was improved through exploration.

\begin{table}[h]
\centering
\begin{tabular}{c | r r | r r} 
  & \multicolumn{2}{c|}{Business KPIs} & \multicolumn{2}{c}{Model metrics} \\
 \textbf{Group} & \textbf{Revenue} & \textbf{CTR} & \textbf{AUC} & \textbf{Log loss} \\
 \hline
 Uncertainty & 5.0\% & 9.6\% & 0.24\% & -0.26\% \\ 
 Random & 3.2\% & -2.3\% & 0.19\% & 0.8\% \\
\end{tabular}
\caption{\textbf{Results of the online A/B test and offline model comparison.} All metrics are presented as relative increases over the control group. Increases are considered to be improvements for all metrics except log loss.}
\vspace{-25pt}
\label{tab:results}
\end{table}

\section{Conclusion}

In this work, we presented a scalable and efficient exploration system integrated into Zemanta's real-time bidding system. The system leverages the uncertainty of deep learning models used for CTR prediction to explore the RTB supply. We use the Monte Carlo dropout algorithm to compute uncertainty of CTR predictions on the request level and direct our exploration into areas where the model's uncertainty is high. Alongside this, we also implemented a controller to automate the exploration. Through online A/B tests, we have shown that this approach brings improvements in business KPIs and model performance.
There is ample opportunity for future work in this area. One direction could be to estimate uncertainty of ads in a similar manner and use both estimates to explore with a more complex strategy, or even combining the two exploration systems into one. We also see great potential in implementing a more sophisticated controller with better insight into our uncertainty estimates that could adjust the level of exploration more precisely.

\section*{Speaker Bio}

\textbf{Jan Hartman} is a data scientist/machine learning engineer at Zemanta, where he works with high-throughput, low-latency machine learning pipelines at a large scale. He leads the initiative for implementing and applying state-of-the-art models for click prediction along with exploration initiatives. Before joining Zemanta, he worked on research projects in the fields of distributed computing, neural network optimization, and cryptography. 
Honors MSc degree in Computer \& Data Science from the University of Ljubljana. Open-source contributor. Research interests include deep learning, neural network embeddings, and model compression. 


\bibliographystyle{ACM-Reference-Format}
\bibliography{bibliography}


\end{document}